\documentclass[sigconf]{acmart}

\usepackage{booktabs} 
\usepackage{subfigure}
\usepackage{cleveref}

\acmDOI{10.1145/nnnnnnn.nnnnnnn}

\acmISBN{978-x-xxxx-xxxx-x/YY/MM}

\acmConference[GECCO '19]{the Genetic and Evolutionary Computation Conference 2019}{July 13--17, 2019}{Prague, Czech Republic}
\acmYear{2019}
\copyrightyear{2019}

\acmPrice{15.00}

\begin{document}


\copyrightyear{2019} 
\acmYear{2019} 
\setcopyright{licensedothergov}
\acmConference[GECCO '19]{Genetic and Evolutionary Computation Conference}{July 13--17, 2019}{Prague, Czech Republic}
\acmBooktitle{Genetic and Evolutionary Computation Conference (GECCO '19), July 13--17, 2019, Prague, Czech Republic}
\acmPrice{15.00}
\acmDOI{10.1145/3321707.3321737}
\acmISBN{978-1-4503-6111-8/19/07}

\title{Classification of EEG Signals using Genetic Programming for Feature Construction}
\author{{\'I}caro Marcelino Miranda}
\affiliation{%
  \institution{University of Bras{\'i}lia}
  \city{Bras{\'i}lia}
  \country{Brazil}
}
\email{icaro.marcelino@hotmail.com}

\author{Claus Aranha}
\affiliation{%
  \institution{Tsukuba University}
  \city{Tsukuba} 
  \country{Japan}}
\email{caranha@cs.tsukuba.ac.jp}

\author{Marcelo Ladeira}
\affiliation{%
  \institution{University of Bras\'ilia}
  \city{Bras\'ilia}
  \country{Brazil}
}
\email{mladeira@unb.br}

\renewcommand{\shortauthors}{\'I. M. Miranda et al.}

\begin{abstract}
The analysis of electroencephalogram (EEG) waves is of critical importance for the diagnosis of sleep disorders, such as sleep apnea and insomnia, besides that, seizures, epilepsy, head injuries, dizziness, headaches and brain tumors. In this context, one important task is the identification of visible structures in the EEG signal, such as sleep spindles and K-complexes. The identification of these structures is usually performed by visual inspection from human experts, a process that can be error prone and susceptible to biases. Therefore there is interest in developing technologies for the automated analysis of EEG. In this paper, we propose a new Genetic Programming (GP) framework for feature construction and dimensionality reduction from EEG signals. We use these features to automatically identify spindles and K-complexes on data from the DREAMS project. Using 5 different classifiers, the set of attributes produced by GP obtained better AUC scores than those obtained from PCA or the full set of attributes. Also, the results obtained from the proposed framework obtained a better balance of Specificity and Recall than other models recently proposed in the literature. Analysis of the features most used by GP also suggested improvements for data acquisition protocols in future EEG examinations.
\end{abstract}

%
%

\ccsdesc[500]{Mathematics of computing~Dimensionality reduction}
\ccsdesc[500]{Computing methodologies~Genetic programming}

\keywords{Classification, EEG, Dimensionality Reduction, Feature Construction, Feature Selection, Genetic Programming, K Complex, Sleep Spindles}
\maketitle

\section{Introduction}

About 40\% of the world's population suffers from some sleep disorder \cite{wade2008prolonged, ohayon2011epidemiological}. Sleep quality directly affects the health and quality of life of the human being. Poor sleep causes many people to seek out specialized clinics for an accurate diagnosis. One of the most common techniques of analysis is done by observing brain activity, eye movement, muscle tension, and other body signals by polysomnography (PSG). The examination consists of collecting data through a series of electrodes connected to the patient's skin and scalp during his or her usual nighttime sleep.
    
This examination allows the diagnosis of several disorders, such as obstructive sleep apnea, insomnia, narcolepsy, restless legs syndrome and bruxism. It is also useful for the identification of visible waveforms like sleep spindles (SS) and K-complexes (KC) which, besides assisting in sleep staging, are related to the consolidation of memory and sensory systems. Abnormalities in their forms may indicate neuropathologies or sleep disorders.

In patients with sleep or neurological disorders, the study of these waveforms helps in the understanding of the neurophysiological functioning and thus, allows to raise hypotheses about the problem. In particular, sleep spindles have a number of theoretical and clinical implications in understanding how brain activity during sleep is affected and the development of the disorder \cite{weiner2016spindle}.
   
\begin{figure}[!b]
    \centering
    \includegraphics[width = .69\linewidth]{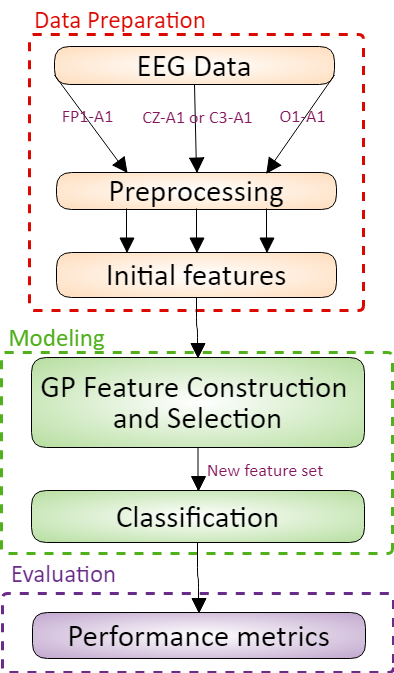}
    \caption{Framework proposed in this work for the identification of structures in sleep EEG}
    \label{fig:flow}
\end{figure}

\begin{figure}[!t]
\center
\subfigure
{\includegraphics[width=.95\linewidth]{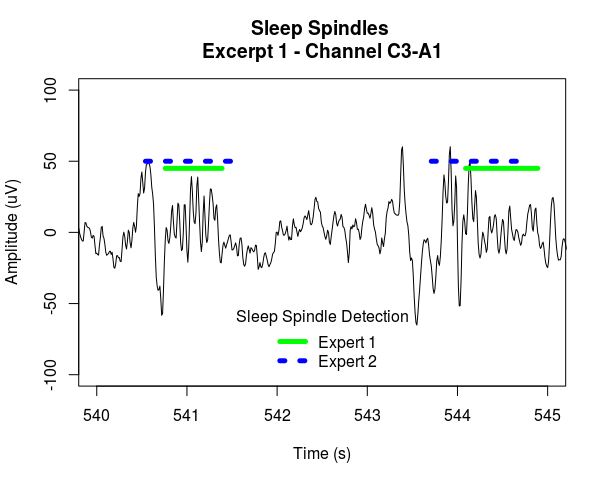}}
\subfigure
{\includegraphics[width=.95\linewidth]{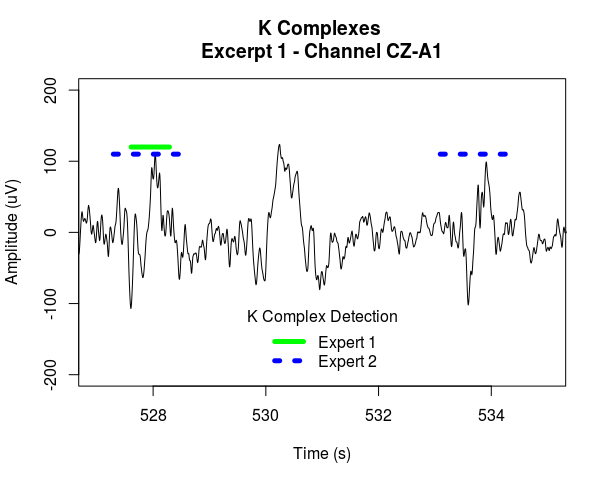}}
\caption{Example of Sleep Spindles and K-complexes identified from EEG data by two different experts.}
\label{ss_and_kc}
\end{figure}

The identification of waveforms on EEG signals is usually done by visual inspection by experts. This is a time-consuming and tiring process, which may introduce biases and errors~\cite{silber2007visual}. In consequence, specialists not always arrive in the same identification, as illustrated in Figure~\ref{ss_and_kc}.
Because of this, there is an interest in the development of automated tools for 
waveform detection~\cite{weiner2016spindle}.

There are many challenges related to EEG signal analysis. They have spatial and temporal co-variance, implying highly dependent samples. They are also non-stationary, noisy and sensitive to external interference~\cite{kevric2017comparison}. 
In order to describe these signals without losing information, a high number of features are necessary from the original signals, implying in high dimensionality samples. 

One way to improve the automated classification of EEG signal structures is by using dimensionality reduction and feature construction techniques. In this sense, Genetic Programming (GP) can be used to generate a function that generates a set of new, reduced features from the original ones. Using GP for feature construction has two advantages: First, GP can generate non-linear combinations of features, making it more expressive than traditional feature reduction techniques. Second, an analysis of the rules created by GP may allow insights about the importance of the different original features, as suggested by Ivert et al.~\cite{ivert2015feature}.


Guo et al. have previously proposed a GP framework for feature construction in the context of seizure detection in EEG signals using KNN~\cite{guo2011automatic}. In this paper, we build upon this work and apply it to the more difficult problem of detecting structures such as Sleep Spindles and K-complexes. More specifically, we use short samples (2s vs 26s) to precisely identify the locations of the structure, we use AUC instead of Precision as the fitness function, and we explore five different classifiers instead of just KNN.

To test the proposed framework (Figure~\ref{fig:flow}) we perform the identification of Sleep Spindles and K-complexes on the DREAMS~\cite{devuyst2011automatic} dataset. Starting from a set of 75 features per sample, the proposed GP finds a constructed set with a median of 12 features. We show that the feature set found by GP achieved better AUC than using the full set of features, or even a set of 29 features selected by PCA. Additionally, the proposed model achieve a better balance of Recall and Specificity when compared with other recently proposed models for the same problem. Finally, and perhaps more interestingly, an analysis of the rules constructed by GP showed that we could use only one of the three EEG channels in the dataset to obtain the same quality of identification. This result suggests that a simpler examination could be used, causing less discomfort for patients.

\section{The EEG Classification Problem}
    
Electroencephalogram (EEG) is a typically noninvasive examination for the observation of electrical activity of the brain \cite{subha2010eeg}. This information is obtained through electrodes attached to the scalp with a conductive paste. Through the analysis of these data it is possible to detect diseases and psychiatric and neurological problems. Usually the analysis of these signals done visually by experts (Figure \ref{ss_and_kc}), which makes the process tiresome, tedious and susceptible to errors~\cite{silber2007visual}. 
    

To assist specialists in this visual task, a number of methods of automatic processing and analysis of EEG signals has been proposed. We emphasize the use of automatic methods for the study of apnea \cite{tagluk2010classification}, epilepsy \cite {guo2011automatic}, drowsiness \cite{da2016automated}, sleep spindles \cite{devuyst2011automatic, devuyst2006automatic, al2018efficient, lachner2018single, yucelbacs2018automatic, d2018different}, K complexes \cite{hernandez2016comparison, ranjan2018fuzzy, sun2018k}, sleep stages \cite{lajnef2015learning} and schizophrenia \cite{roschke1995nonlinear}.


\subsection {Sleep Spindles and K-complexes}
Sleep has two main phases: REM (Rapid Eye Movement) sleep and NREM (non-REM) sleep. Occupying up to 80\% of the sleep time, the NREM phase is divided into 4 stages, ranging from lightest to deepest sleep \cite{rama2006normal}.

In particular, stage 2 of NREM sleep has as its main characteristic the appearance of specific waveforms, K-complexes and Sleep Spindles. The beginning of this stage is defined by the occurrence of these signals. Because of this well defined presence, they are very important for sleep staging.

Although Sleep Spindles mark entry into stage 2 of NREM sleep, they may also occur in stage 3 \cite{berry2012aasm}. When a spindle occurs, the amplitude of the EEG signal increases and decreases progressively, having a minimum duration of 0.5 s with defined bandwidth between 12 and 14 Hz in the criterion of Rechtschafen and Kales \cite{rechtschaffen1968manual} (some authors may consider from 11 until 16 Hz). Peak-to-peak amplitude settings can also be found between 5 and 25 $\mu V$ \cite{devuyst2011automatic}. The occurrence of spindles contributes to memory consolidation, to continuous sleep \cite{clemens2005overnight} and in the study of sleep and neurological disorders.

The characteristics of the spindles change with the patient's age and sex \cite{de2003sleep}. The tendency is for it to occur less with advancing age \cite{peters2014age}. As for gender, the phenomenon usually occurs twice more during the sleep of women, due to hormonal factors \cite{manber1999sex, dzaja2005women}.

The sleep spindles, in general, have a well-defined structure (occurrence, bandwidth, and amplitude). However the advancement of the patient's age and pathologies cause inaccuracies in their shape. Typically, the number of spindles decreases and their shape deteriorates \cite{jose1982sleep}. Their shape may be distorted and are more subject to the occurrence of interference \cite{devuyst2006automatic}. Patients with schizophrenia, for example, do not have normal patterns in the spindles \cite{ferrarelli2007reduced}. Changes can also be observed due to fatal familial insomnia \cite{niedermeyer2000considerations}, autism and epilepsy \cite{iranmanesh2017ultralow}. This lack of standard is important for the diagnosis of neurological diseases, but it makes it more difficult to identify this phenomenon for specialists and automatic methods.

The K-complex is a negative acute wave immediately followed by a positive component that clearly arises in the EEG, having a minimum duration of 0.5s in the frequencies of 12 to 14 Hz \cite{berry2012aasm}. In the identification, they can be easily confused with any waveform with high peaks~\cite{krohne2014detection, erdamar2012wavelet} (Figure \ref{ss_and_kc}). Abnormal activity of the K complex may be related to epilepsy, restless leg syndrome, and obstructive sleep apnea.
 
\subsection{DREAMS Data}
We use the databases collected by the DREAMS project~\cite{devuyst2011automatic}, which consist of a series of polysomnography (PSG) with expert annotations on phenomena or sleep disorders. We use the sleep spindle and K-complexes datasets from this project. Their purpose is to tune, train and test automatic detection algorithms.

The Spindles dataset consists of 30 minute stretches of the central EEG channel (extracted from full-night PSG records), independently annotated by two experts. The data were acquired in a sleep laboratory of a Belgian hospital (BrainnetTM System of MEDATEC, Brussels) using a 32-channel digital polygraph. It is important to highlight that all records on this dataset are from patients with various sleep pathologies: dyssonia, restless legs syndrome, insomnia, apnea syndrome or hypopnea. EOG, EMG and EEG channels (channels FP1-A1, O1-A1 and C3-A1 or CZ-A1) were recorded, using the European standard data format (EDF) for storage. Only EEG channels will be used in this work. 


The sampling frequency varies between patients, having records of 200Hz, 100Hz or 50Hz of 30 minutes duration. 
The recordings were given independently to two experts, who annotated their estimates for the locations of the sleep spindles. 
The K-complex records were collected in the same hospital as the spindle registers, with the same equipment. There are 10 polysomnographic recordings from healthy individuals. Just like the previous base, we only use EEG channels. The sampling frequency was 200 Hz for all patients with a 30 min duration. In the same way, the excerpts were given independently to two experts. To reduce bias, the experts did not have access to sleep staging of the records.

\subsection{Data preparation}
The original EEG data was prepared for the automated identification process using the following procedure.

\subsubsection{Wavelets Transform}
Wavelet transforms are mathematical tools capable of decomposing signals into several components that allow analysis at different time and frequency scales \cite{de2002wavelets}.

\begin{figure}[!htb]
         \centering
        \includegraphics[width = \linewidth]{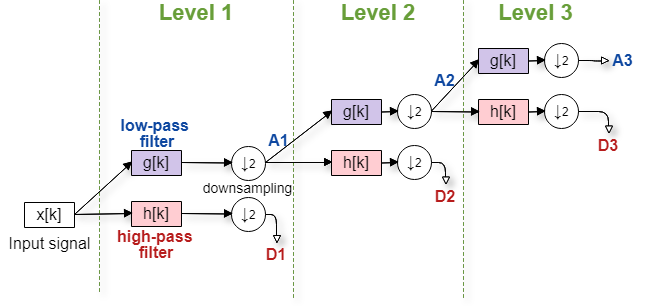}
        \caption{Example of Digital Wavelet Transform in 3 levels}
        \label{fig:wavelets}
\end{figure}

 An input signal $x$, passes through a low-pass filter $g$ and a high-pass filter $h$ (parallel, not sequentially), each with a cut-off frequency equal to one half of the sampling frequency of the input. Then, the two generated sub-signals, that is, the output of the filters has their samples reduced in half (see Figure~\ref{fig:wavelets}).
 
 This process can be repeated at several levels, causing the output of the low pass filter to be the input signal of a new pair of filters, followed by downsampling.
 
\subsubsection{Feature extraction}

The EEG data used were sampled with different frequencies (50, 100 or 200 Hz). For the application of the Digital Wavelet Transform (DWT) \cite{mallat1989theory} with 5 decomposition levels, the data were resampled with an increased frequency of 256 Hz in all cases by means of interpolation through a cubic spline.
    
For each decomposition level ($D1$ to $D5$) in each EEG channel, the signal was separated into 2 second samples, and the following attributes were calculated for each sample: Signal amplitude average, Signal amplitude standard deviation (SD), Symmetry, Power Spectral Density (PSD), and Signal curve length.
    
Following this procedure, we obtain 900 samples (2s samples from 30 minutes of signal) with 75 real valued attributes (3 EEG channels, 5 DWT levels, and 5 attributes per level). These attributes are summarized in tables \ref{tab:att_eeg_c}, \ref{tab:att_eeg_fp}, and \ref{tab:att_eeg_o}. Here, the columns represent the coefficients of the DWT levels and the lines the operations performed.


\begin{table}[!htb]
    \centering
    \begin{tabular}{c|c c c c c}
                & D1 & D2 & D3 & D4 & D5 \\\hline
      Average   & ARG0 & ARG5 & ARG10 & ARG15 & ARG20 \\
      SD & ARG1 & ARG6 & ARG11 & ARG16 & ARG21 \\
      Symmetry  & ARG2 & ARG7 & ARG12 & ARG17 & ARG22 \\
      PSD       & ARG3 & ARG8 & ARG13 & ARG18 & ARG23 \\
      Curve Length& ARG4 & ARG9 & ARG14 & ARG19 & ARG24 \\
    \end{tabular}
    \caption{Attributes for the central EEG channel (CZ-A1 or C3-A1)}
    \label{tab:att_eeg_c}
\end{table}

\begin{table}[!htb]
    \centering
    \begin{tabular}{c|c c c c c}
                    & D1 & D2 & D3 & D4 & D5 \\\hline
      Average & ARG25 & ARG30 & ARG35 & ARG40 & ARG45 \\
      SD & ARG26 & ARG31 & ARG36 & ARG41 & ARG46 \\
      Symmetry & ARG27 & ARG32 & ARG37 & ARG42 & ARG47 \\
      PSD & ARG28 & ARG33 & ARG38 & ARG43 & ARG48 \\
      Comp. Length&ARG29 & ARG34 & ARG39 & ARG44 & ARG49 \\
    \end{tabular}
    \caption{Attributes for the EEG channel FP1-A1}
    \label{tab:att_eeg_fp}
\end{table}

\begin{table}[!htb]
    \centering
    \begin{tabular}{c|c c c c c}
                    & D1 & D2 & D3 & D4 & D5 \\\hline
      Average & ARG50 & ARG55 & ARG60 & ARG65 & ARG70 \\
      SD      & ARG51 & ARG56 & ARG61 & ARG66 & ARG71 \\
      Symmetry& ARG52 & ARG57 & ARG62 & ARG67 & ARG72 \\
      PSD     & ARG53 & ARG58 & ARG63 & ARG68 & ARG73 \\
      Curve Length & ARG54 & ARG59 & ARG64 & ARG69 & ARG74 \\
    \end{tabular}
    \caption{Attributes for the EEG channel O1-A1}
    \label{tab:att_eeg_o}
\end{table}

\section{Proposed Framework}

Previously, Guo et al.~\cite{guo2011automatic} proposed the use of Genetic Programming (GP) for the construction of  features for EEG analysis in the classification of epileptic episodes. Taking this framework as a base, we develop a framework for the identification of structures in sleep EEG, which we describe in this section.

There are many characteristics in the structure identification problem which differentiates it from the earlier classification work. We must divide the EEG signals into multiple short samples in order to identify the position of the Spindles and K-complexes in the signal. As a consequence, the data becomes highly unbalanced, complicating the training process. 
Also, we work on three distinct EEG channels (as opposed to a single channel in the original work). 

We tested several improvements on the original work to deal with this harder problem. First, we use the Area Under the ROC Curve (AUC) of the classifiers instead of the accuracy as the fitness measure, since the AUC is more robust and discriminating~\cite{huang2005using}. Also, we compare several classifiers in addition to KNN, to explore the relationship between classifier choice and GP feature construction. 

GP is widely applied in the construction and selection of features for its good performance. In classification problems it is possible to evolve a tree for each problem class, selecting  the best attributes and creating new features for each of them \cite{muni2006genetic}. Even with unbalanced data, this approach with GP also gets good results \cite{viegas2018genetic}. In literature, there are also application studies in benchmarks \cite{suarez2014genetic} and in databases with few samples \cite{nandi2006classification}.

Finally, we publish the program of the proposed framework and experiments at our repository\footnote{\url{https://github.com/IcaroMarcelino/SleepEEG}} for reproducibility purposes. 

\subsection{GP for Feature Construction}

Our proposed framework uses Genetic Programming (GP) to generate the constructed set of features from the original features. 

The GP tree is defined as follows: The input nodes are selected from the original attributes. The intermediate nodes are selected from a set of arithmetic operators $\{+,-,\times,\}$, as well as a set of \emph{protected} operators $\{/,\ln,\sqrt\}$. These protected operators have their definitions slightly modified to avoid errors such as division by zero, as follows:

\begin{eqnarray*}
    \text{protected division} (a, b) &=& \left \{
        \begin{array}{l l}
            1, & b = 0 \\
            \frac{a}{b}, & b \neq 0
        \end{array} \right.\\
    \text{protected log}(a) &=& \left \{
        \begin{array}{l l}
            1, & b = 0 \\
            ln(|a|), & b \neq 0
        \end{array} \right.\\
    \text{protected square root} (a) &=& \sqrt{|a|}
\end{eqnarray*}

Additionally, a special "Feature Operator" \cite{guo2011automatic}, $F$, is used to indicate how to obtain the constructed features from the GP tree. The feature operator returns the value of its input as its output, without any changes. Its purpose is to mark a subtree as one of the constructed features. 
    
Each $F$ operator will be the root of a subtree that expresses the function to calculate one attribute for the constructed set. In this way, a GP tree containing ten nodes with the $F$ operator will generate a constructed attribute set with 10 attributes. 
    
\begin{figure}[!htb]
   \centering
    \includegraphics[width=.45\linewidth]{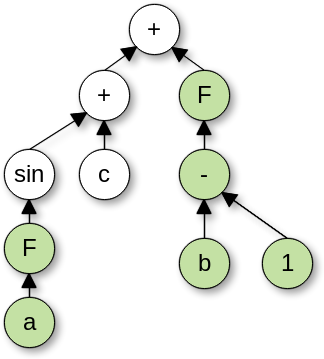}
    \caption{Example of GP program marked with F operators. This GP tree constructs two attributes: $F_1 = a$ and $F_2 = b-1$}
    \label{fig:ex_F}
\end{figure}

For example, the tree depicted in Figure~\ref{fig:ex_F} shows a GP individual with two subtrees marked by the $F$ operator. If we assume that the original attribute set has 26 attributes (a..z), this tree will generate a constructed attribute set with two attributes: $F_1 = a$ and $F_2 = b - 1$. 
    
The use of the $F$ operator allows a single GP tree to express multiple attribute constructing functions. In this way, we avoid having to explicitly define how many attributes will be constructed beforehand, which would be necessary if each attribute were expressed by a separate tree~\cite{guo2011automatic}. 

Also, this allows GP trees in the population to exchange useful subtrees containing sets of constructed attributes that were successful. We believe that this allows the GP to pass around the most relevant subtrees to the next generations and, with this, keep attributes and attribute subsets that facilitate classification.
    
To evaluate one GP tree using the structure described in the previous section, we use the following procedure. First, we generate the set of constructed attributes for the GP tree using the $F$ operators. Second, we train a classifier using this set of constructed attributes. Finally, we use the AUC value of the classifier as the fitness value for the GP tree. In this way, the product of the evolutionary process is both a set of constructed attributes, as well as a classifier trained on those attributes. The parameters used for the evolutionary process in the current framework are listed in Table~\ref{tab:gp_param}. As the parameters provided good results in the initial runs, they were maintained for the following. The "Uniform mutation" selects a random node from the individual and replaces subtree rooted in that node with a a randomly generated one.
    
\begin{table}[!htb]
    \centering
    \begin{tabular}{r|l}
    Parameter & Value \\\hline
    Number of generations & 300 \\
    Number of individuals & 100 \\
    Selection & Tournament (size = 3) \\
    Crossover & One Point \\
    Mutation & Uniform \\
    Primitive Functions & $+, -, /, *, ln, \sqrt{}, F$\\
    Fitness & AUC
    \end{tabular}
    \caption{GP Model Parameterization}
    \label{tab:gp_param}
\end{table}

\subsection{Classifiers}
    
We compare the performance of the set of constructed attributes by GP by using five different classifiers:
\begin{itemize}
    \item K Nearest Neighbors (KNN)
    \item Naive Bayes (NB)
    \item Support Vector Machines (SVM)
    \item Decision Tree (DT)
\item Multilayer Perceptron (MLP)
\end{itemize}

We perform an initial tuning procedure using the full set of 75 features to select the hyperparameters used by each classifier in the next experiments. For each tuning classifier, we execute 100 runs for each parameter value tested in the following sets: 
\begin{itemize}
    \item KNN: $k \in \{3,5,7,9,11,13,15,17,19\}$
    \item SVM: kernel $\in \{$Radial Basis Function (RBF), polynomial, sigmoid$\}$
    \item MLP: activation function $\in \{$ReLU, logistic$\}$, neurons in hidden layer $\in \{15,30,45,60,75\}$
\end{itemize}

The parameters that provide the best performance were selected. KNN with $k = 5$, SVM with RBF as kernel and MLP with a single hidden layer with 15 neurons and ReLU activation.

\section{Experiment}

 We perform several evaluations of the classifiers on the Sleep Spindles and K-Complexes datasets in order to analyse the performance of the propose framework. The results of the classifiers are compared using the full set of 75 attributes, a reduced set of 29 attributes selected by PCA, and the set of attributes constructed by the GP framework.

The training dataset (used to train both the GP and the classifiers) was generated by simple random sampling 70\% of the signal samples, and labelling them as positive samples (Sleep Spindles or K-complexes) if both specialists agreed on the label. Additionally, because the dataset is highly unbalanced, we balance the training dataset by randomly removing samples from the majority class until both classes have the same number of signal samples.

The test dataset was generated by the remaining 30\% of the samples, and each signal sample was labeled as positive if either specialist annotated it as positive. Also, the balancing procedure is not performed on the testing data set. This resulted in a slightly harder testing data set.

For each experiment, we repeat the training/testing procedure 10 times, and report the aggregate results of these 10 repetitions as described in the subsections below.


\begin{figure*}[!htb]
\center
\subfigure[refA][Classifiers performance over the 75 initial attributes]{\includegraphics[width=.48\linewidth]{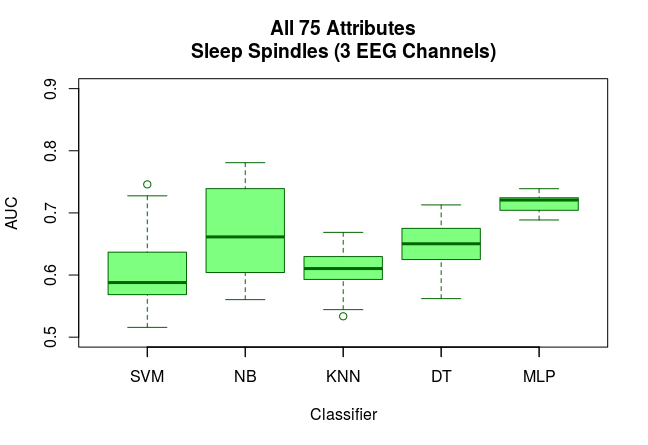}}
\qquad
\subfigure[refB][Classifiers performance over 29 PCA components]{\includegraphics[width=.48\linewidth]{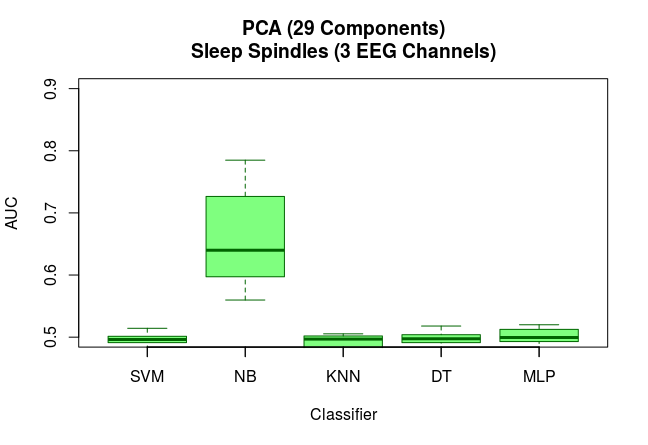}}
\subfigure[refC][Classifiers performance over reduced feature sets generated by GP]{\includegraphics[width=.48\linewidth]{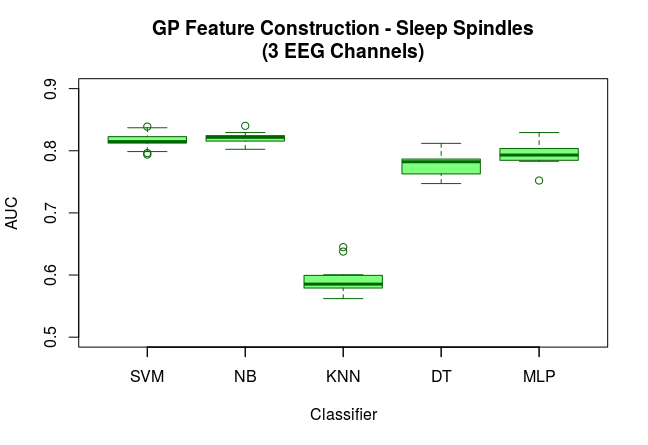}}
\qquad
\subfigure[refD][Classifiers performance over reduced feature sets generated by GP]{\includegraphics[width=.48\linewidth]{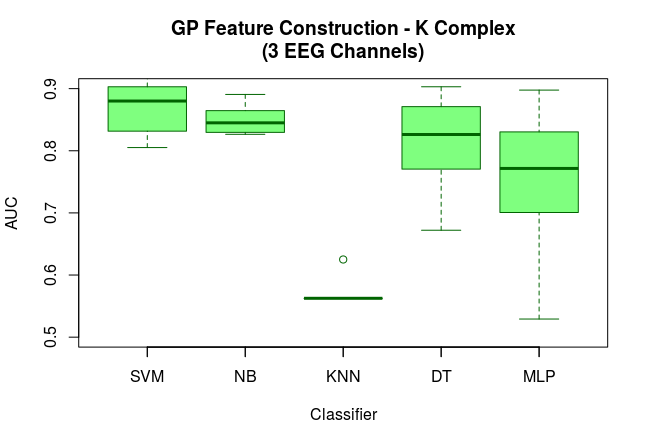}}
\subfigure[refE][Classifiers performance over reduced feature sets generated by GP (Male Patients)]{\includegraphics[width=.48\linewidth]{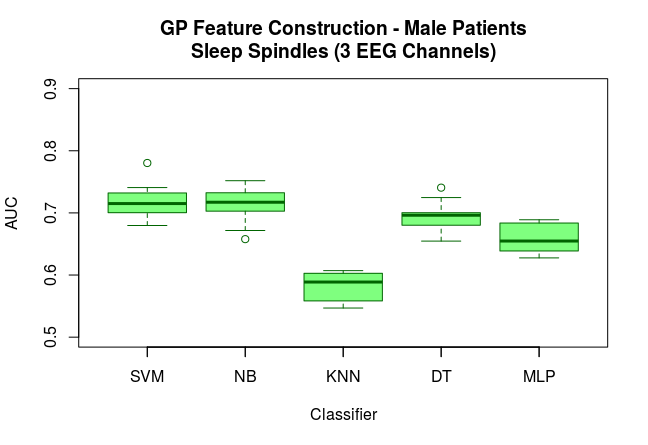}}
\qquad
\subfigure[refF][Classifiers performance over reduced feature sets generated by GP (Female Patients)]{\includegraphics[width=.48\linewidth]{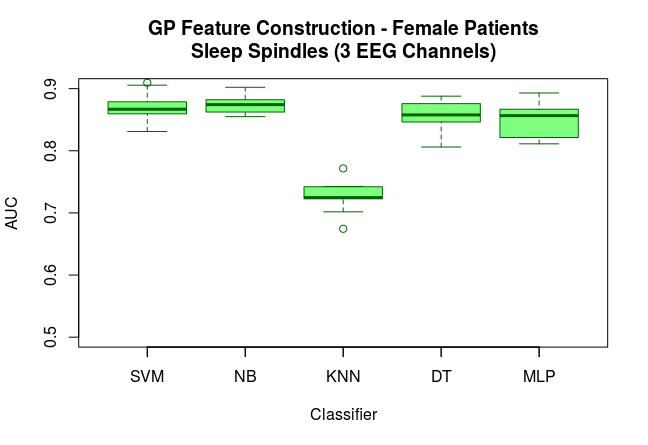}}
\caption{Performance of the Classifiers on the test set. a- full attribute set (Sleep Spindles), b- PCA attribute set (Sleep Spindles), c- GP attribute set (Sleep Spindles), d- GP attribute set (K-complex), e- GP Attribute (Sleep Spindles, Males only), f- GP Attribute (Sleep Spindles, Females only)}
\label{results}
\end{figure*}

\subsection{Results}
The first experiment was aimed at verifying the performance of the classifiers on the test dataset without the reduction of dimensionality by GP, i.e., using the 75 features. The results are also useful to justify the application of feature construction. 


The classifiers performance can be seen in Figure~\ref{results}a. Only MLP has good results, achieving an AUC greater than 0.7 with low SD. 

    \begin{figure*}[!htb]
         \centering
         \includegraphics[width = \linewidth]{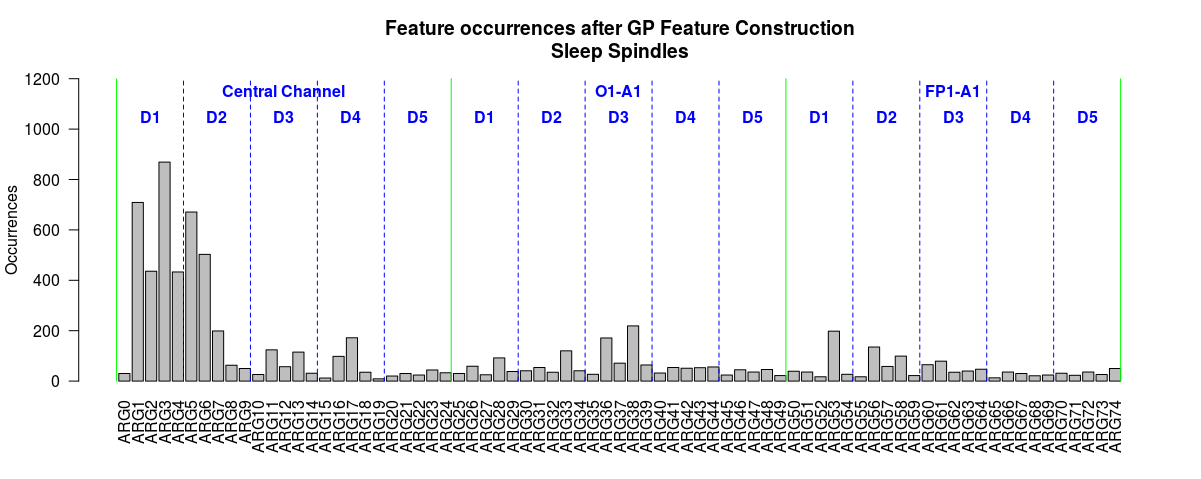}
         \caption{Number of occurrences of the feature in the models generated by GP (Same models from Figure \ref{results}c).}
         \label{fig:feat_freq}
    \end{figure*}

    \begin{figure}[!htb]
         \centering
         \includegraphics[width = \linewidth]{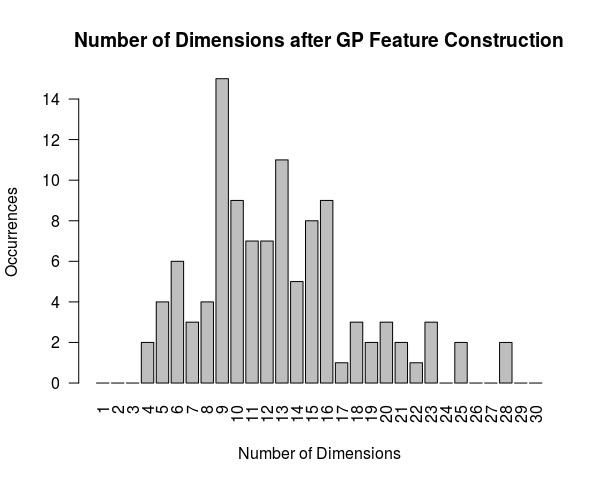}
         \caption{Number of dimensions in the models generated by GP (Same models from Figure \ref{results}c)}
         \label{fig:dim_freq}
    \end{figure}

Applying PCA on the data with a 95\% threshold for variance, ensuring little loss of information, the initial set of 75 features can be represented with 29 attributes.

    
The performance of the trained classifiers with the feature set generated by PCA on the respective test set can be seen in Figure~\ref{results}b. For this problem, the PCA representation caused a performance reduction in all classifiers except NB.

The performance of applying the classifiers on the feature set generated by PCA can be seen in Figure~\ref{results}b. For this problem, the PCA representation caused a decrease in performance in all classifiers except NB.

Applying the GP feature reduction, the AUC of the classification increases for all classifiers except KNN (Figure~\ref{results}c) and the SD reduces for all cases. This method is capable to reduce the number of features from 75 to less than 29 (Figure~\ref{fig:dim_freq}).

Using the same approach for training K-complexes classifiers, the models achieve high AUC scores too (Figure~\ref{results}d). 

\begin{figure}[!htb]
\center
\subfigure{\includegraphics[width=\linewidth]{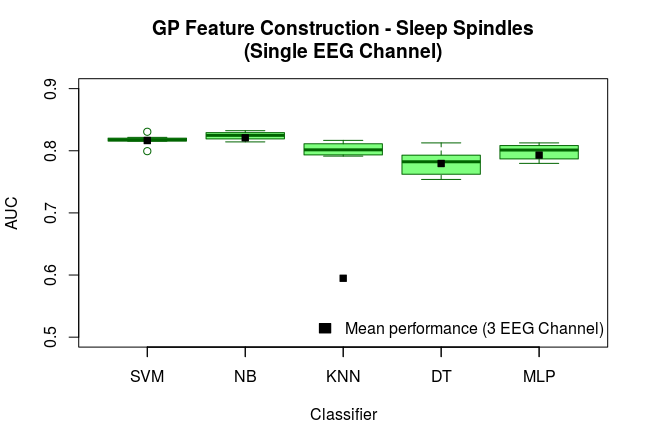}}
\subfigure{\includegraphics[width=\linewidth]{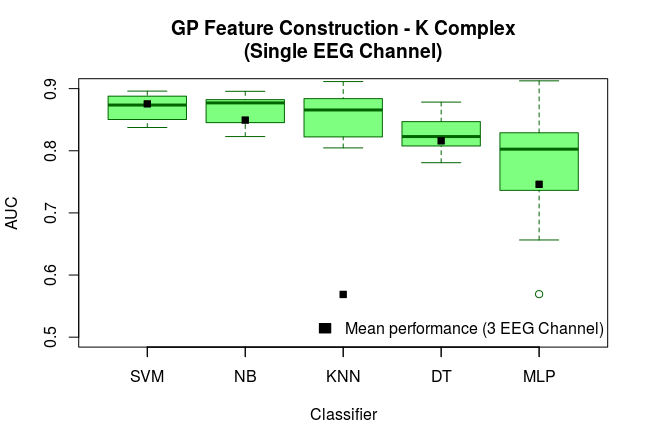}}
\caption{Classifiers performance over the central EEG channel attributes for Sleep Spindles and K-complexes}
\label{fig:central}
\end{figure}

\subsection{Analysis of gender difference}
 \begin{table}[!htb]
    \centering
    \begin{tabular}{c|c|c}
	    Gender  &  Expert 1     & Expert 2 \\\hline
        Male    &  157          & 121      \\
        Female  &  198          & 288       
	    \end{tabular}
    \caption{Spindles scored by the experts for each gender}
    \label{tab:gender}
\end{table}
As mentioned before, there are gender differences in sleep spindles. To see if this difference affects the performance of the classifiers, the data were separated by gender. 

Observing the Figures \ref{results}e and \ref{results}f, the classifier's performance for female patients is higher than the male patients. As sleep spindles occur more often in female patients, in data there are more representative samples of the waveform (see Table \ref{tab:gender}), which facilitates the training of more efficient classifiers. 


\subsection{Constructed Features Analysis}
In Figure~\ref{fig:feat_freq}, the frequency of occurrence of features in the models training shows that there are attributes more relevant than others in the dataset. The greater occurrence of the features associated to the central EEG channel indicates that it has more important role to the identification of sleep spindles. 


Using only this channel for training (i.e, only with the 25 first features of the dataset), the performance of all classifiers increase for sleep spindles and K-complexes identification (Figure~\ref{fig:central}). This attributes reduction contributes to a better understanding of the phenomenon, providing a more efficient approach  by reducing the use of electrodes, consuming less resources and generating less discomfort to the patient.

 \begin{table}[!htb]
    \centering
    \begin{tabular}{c|c|c|c|c}
	    Reference & Recall  & Specif. & Prec. & $F_1$  \\\hline
        Lachner-Piza et al., 2018 \cite{lachner2018single} & 0.65 & 0.98 & 0.38 & 0.48 \\
        Tsanas and Clifford, 2015 \cite{10.3389/fnhum.2015.00181}  & \textbf{0.76} & 0.92 & 0.33 & 0.46 \\
        Zhuang et al., 2016 \cite{Zhuang2016}  & 0.51 & \textbf{0.99} & \textbf{0.70} & \textbf{0.59} \\
        \textbf{Proposed model}   & \textbf{0.75} & \textbf{0.98} & 0.35 & 0.48\\
	    \end{tabular}
    \caption{Comparison between the proposed model and literature models}
    \label{tab:compare}
\end{table}

\section{Comparison to Literature Models}
In the Table \ref{tab:compare}, the best generated model with the proposed approach (with NB classifier) was compared with the literature models which also used DREAMS data \footnote{Further comparisons between sleep spindle identifiers can be seen in \cite{lachner2018single}}.

Tsanas at. el \cite{10.3389/fnhum.2015.00181} and Zhuang et al. \cite{Zhuang2016} proposed continuous wavelet transform (CWT) based approaches and the estimation of the probability of spindles occurrences. Lachner-Piza et al. \cite{lachner2018single} proposed a SVM approach with a feature selection method based on the label-feature and feature-feature correlations for determining the relevance and redundancy of each feature.

Observing the performance of the models, all obtained high specificity, indicating that the identification of samples where no spindles samples are present is reliable.
Moreover, there is a trade-off between sensitivity and precision. In the context of applying automatic identifiers, false negatives are more unwanted than false positives. That is, a highly accurate but not very sensitive classifier generates many false positives, indicating that it is not judicious. In a semi-automatic application with low sensitivity, it is necessary for a specialist to inspect the markings performed by the classifier, eliminating the excess of false positives. This is the case of the detector of Zhuang et al. \cite{Zhuang2016}.

The detectors of Tsanas and Clifford  \cite{10.3389/fnhum.2015.00181} and Lachner-Piza et al. \cite{lachner2018single} and the proposed model have achieved a better compromise between sensitivity (recall) and precision. This implies that the identification of signal stretches as spindles is more reliable.

The proposed approach allows the generation of competitive models with the literature. The Tsanas and Clifford model, although having a slightly higher sensitivity than the proposed model. In contrast to the model of Lachner-Piza et al., our model has only minor precision, with 0.03 of difference.

    


\section{Discussion}
The GP feature construction improves the performance of a classifier reducing the search space and generating more explicit relations between variables. Observing the reduction in the number of attributes, the dimensionality of the problem is reduced by up to 7 times in most cases. In addition, by analyzing the most frequent attributes, it is clear which ones are most relevant to the models. With this information, it is possible to select the most important EEG channels.

In the case of K-complexes sleep spindles, only the central EEG channel is sufficient to perform the waveform identification. The single channel approach already reduces search space by one-third. Furthermore, fewer electrodes will be required for the examination, making it more comfortable for the patient, consequently, approaching the sleep in the laboratory of the daily sleep, avoiding biases.

\section{Conclusions}

    The use of automatic methods to identify sleep phenomena makes it possible to classify EEG signal segments with good performance, indicating whether or not a particular event occurs.
    
    Excerpts of 30 minutes can have hundreds of events that you want to identify. In this respect, the proposed model can be used to accelerate the process, and it is up to the expert to assign the classification. The model was also useful for sleep staging, since the presence of spindles and K-complexes strongly characterize sleep stage 2.
    
    It has also been shown that it is possible to significantly improve the performance of classifiers by selecting and constructing attributes. In addition, the use of GP allows greater interpretability and mathematical analysis of the new attributes generated, which may help to better understand the model and the problem. It is also possible to inspect the attributes generated through knowledge in the application domain.
    
    The approach also does not require in-depth knowledge of the application domain. In the first experiment, no  assumptions about the data were performed.
    
    The ease of defining the terms in which the solutions will be written, that is, the operators and the terminals, allow the creation of hybrid models with biomedical information. It can also facilitate communication between specialists from different areas.
    
    The automation of the selection and construction of attributes generates a dataset suitable for the desired classifier. But, it is very simple to apply the attributes generated in another classifier.
    The processing time is also reduced with the smaller number of dimensions.
    
    The PSG generated signals used in sleep clinics are stored directly on computers. Therefore, the application of the proposed technique can be easily applied in this context. The generated models, once trained, make predictions quickly, facilitating a real-time approach.
    
    Measurement of the micro-event activity on EEG signals in different populations can provide important information about abnormalities in brain signals and assist in the investigation and hypothesis assessment of observed phenomena or disturbances. This underscores the importance of the study.
     
    With the proposed models, the identification of the spindles or K-complexes is less costly for the specialist. It may even replace its function in this task if the performance of the models is satisfactory for the requested analysis. Therefore, the diagnosis can be faster and the return to the patient suffering from some disorder is more efficient. Moreover, this methodology can easily be extended to other classification problems.

    \clearpage
    \newpage
    
    



\bibliographystyle{ACM-Reference-Format}
\bibliography{sample-bibliography} 

\end{document}